\documentclass[a4paper, 10pt, conference]{ieeeconf}      
\usepackage{FG2026}
\usepackage{graphicx}
\usepackage{makecell}

\usepackage{graphicx}
\usepackage{amsmath}
\usepackage{amssymb}
\usepackage{booktabs}
\usepackage{subcaption}

\def\B#1{\textbf{#1}}
\usepackage{array}
\newcolumntype{P}[1]{>{\centering\arraybackslash}m{#1}}

\FGfinalcopy 

\IEEEoverridecommandlockouts                              
\def\FGPaperID{38} 

\title{\LARGE \bf
Inter-Stance: A Dyadic Multimodal Corpus for Conversational Stance Analysis
}

\author{\parbox{16cm}{\centering
    {\large Xiang Zhang$^1$, Xiaotian Li$^1$, Taoyue Wang$^1$, Nan Bi$^1$, Xin Zhou$^1$, Cody Zhou$^1$, Zoie Wang$^2$, Andrew Yang$^3$, Yuming Su$^4$, Jeff Cohn$^5$, Qiang Ji$^3$ and Lijun Yin$^1$}\\
    {\normalsize
    $^1$ State University of New York at Binghamton $^2$ Choate Rosemary Hall \\
    $^3$ Rensselaer Polytechnic Institute $^4$ Ward Melville High School $^5$ University of Pittsburgh}}
    \thanks{979-8-3315-7231-0/26/\$31.00 \copyright2026 IEEE}
}

\usepackage{fancyhdr}
\begin{document}
\ifFGfinal
\thispagestyle{empty}
\pagestyle{empty}
\else
\author{Anonymous FG2026 submission\\ Paper ID \FGPaperID \\}
\pagestyle{plain}
\fi
\maketitle
\thispagestyle{fancy}

\begin{abstract}

Social interactions dominate our perceptions of the world and shape our daily behavior by attaching social meaning to acts as simple and spontaneous as gestures, facial expressions, voice, and speech. People mimic and otherwise respond to each other’s postures, facial expressions, mannerisms, and other verbal and nonverbal behavior, and form appraisals or evaluations in the process. Yet, no publicly-available dataset includes multimodal recordings and self-report measures of multiple persons in social interaction. Dyadic recordings and annotation are lacking.  We present a new data corpus of multimodal dyadic interaction  (45 dyads, 90 persons) that includes synchronized multi-modality behavior (2D face video, 3D face geometry, thermal spectrum dynamics, voice and speech behavior, physiology (PPG, EDA, heart-rate, blood pressure, and respiration), and self-reported affect of all participants in a communicative interaction scenario. Two types of dyads are included: persons with shared past history and strangers. Annotations include social signals, agreement, disagreement, and neutral stance. With a potent emotion induction, these multimodal data will enable novel modeling of multimodal interpersonal behavior.  We present extensive experiments to evaluate multimodal dyadic communication of dyads with and without interpersonal history, and their affect. This new database will make multimodal modeling of social interaction never possible before. The dataset includes 20TB of multimodal data to share with the research community.

\end{abstract}

\section{Introduction}

Social signals are relational attitudes that express, often unconsciously, our actual feelings toward interactions and social contexts. Examples include interest, empathy, hostility, agreement, disagreement, flirting, dominance, superiority, etc. 
   Research shows that the dyadic effect exists during interpersonal communication~\cite{jovanovic2006corpus,boker2011real,boker2011something,elfenbein2006brief,barry2000dyadic, chartrand1999chameleon}. 
   People mimic postures, facial expressions, mannerisms and other verbal and nonverbal expressions of the counterpart in social interaction (e.g., contagious effects of laughter and yawing, mimicry of speech rate and rhythms, etc.)~\cite{delaherche2012interpersonal,gatica2009automatic,gueguen2009mimicry,hess2013emotional,lakin2003chameleon}. Interpersonal emotion influence could be reflected on facial and body gesture mimicking, attitude change, and imitation of mannerisms in terms of the underlying affective state (e.g., sadness, empathy, trust, agreement, etc.). Research in psychology and social science has shown that the presence of mimicry behavior can serve as an indicator of co-operativeness, social judgement, etc. Coordinated interpersonal timing, for instance, is related to feelings of rapport; its absence can index a range of disorders (e.g., depression)~\cite{girard2013social,girard2014carma}.

On the other hand, complex human behavior in various social interactions can be better understood by integrating physical features from multiple communicative modalities (e.g., facial expression, speech and vocal expression, gaze, physiological responses, etc.) Although emotions are exhibited in multiple modalities, currently, emotions have been studied mostly in individual modalities. 
Moreover, most existing emotion-related research was mainly based on monadic data collection of an individual with less consideration of dyadic mutual emotion influences~\cite{elfenbein2006brief,barry2000dyadic}. If dual data are collected during dyadic communication, we are able to examine not only individual differences in expressing or communicating emotions but also the dyadic level of analysis with social signal processing. To the best of our knowledge, there is no database of emotional behavior that combines dyadic emotion data with multiple emotion-related modalities: 3D and 2D facial visual dynamics, speech and vocal dynamics, skin temperature dynamics, and physiological responses from two communicating individuals. Therefore, it is highly demanded for creating such a database from multiple modalities in a social interaction setting.

\begin{figure}[t]
    \centering
    \includegraphics[width=0.95\linewidth]{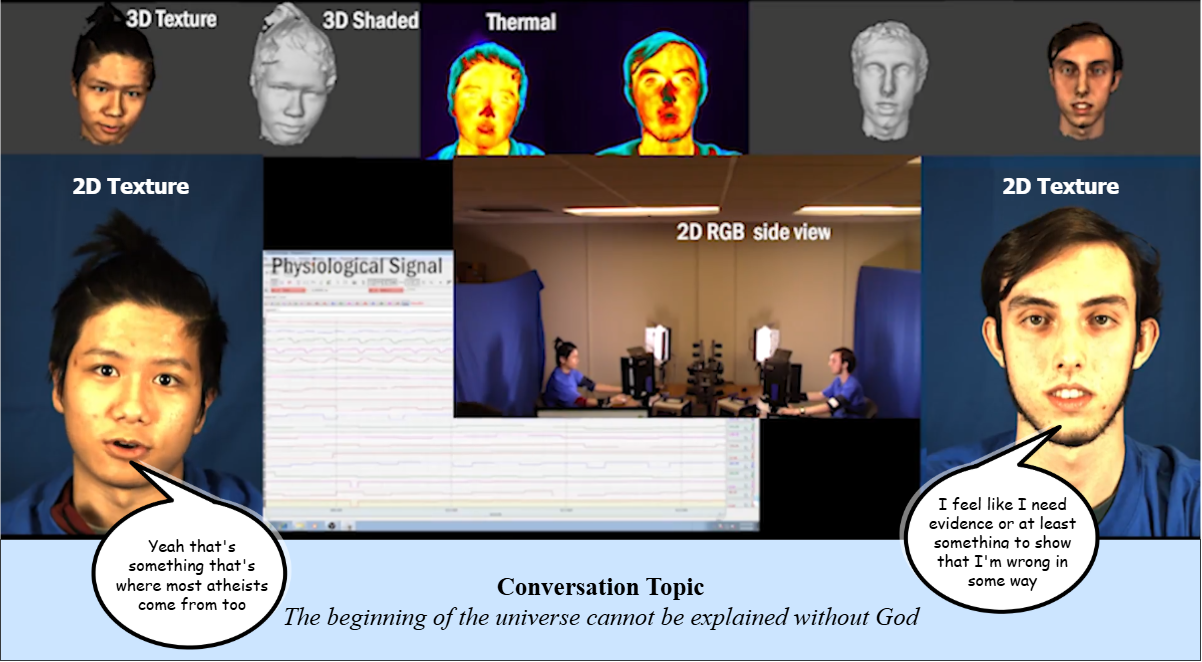}
    \caption{One sample on our Dyadic Multimodal Corpus}
    \label{fig:dyadic_intro}
    \vspace{-5mm}
\end{figure}


Within this landscape, a central challenge for computational analysis is the modeling of \textit{interpersonal stance}, specifically, the visual and physiological expressions, and social signals of agreement, disagreement, and neutrality in face-to-face conversations. Automatically identifying a person's stance has far-reaching applications, from enhancing media monitoring and combating the spread of fake news to developing more socially aware artificial intelligence. 
Stance is inherently multimodal and far more nuanced than simple sentiment polarity; it is a targeted reaction that is deeply embedded in the conversational context. While speech conveys part of the message, the most salient and often most honest cues are nonverbal. These signals range from overt, easily recognizable behaviors such as head nods for agreement and head shakes for disagreement, to a rich set of more subtle indicators. These include fleeting micro-expressions that can leak concealed emotions, minute shifts in 3D facial geometry that betray internal states, thermal patterns linked to physiological arousal, and direct physiological signals measurable by wearable devices.


Is it worth noting that the data acquisition problem is compounded by the fact that existing resources force a trade-off between modality richness and interactive realism. 
For example, the state-of-the-art BP4D+ and BP4D++ datasets~\cite{bp4d+_2016, bp4d++2023}, offer an exemplary suite of sensors synchronized 2D video, 3D facial geometry, thermal imaging, and physiological measures, but it is limited to unilateral capture of single-subjects, precluding any study of dyadic dynamics. Conversely, interaction-focused datasets such as IEMOCAP~\cite{busso2008iemocap} provide invaluable audiovisual and motion capture data from pairs of interacting individuals, but they lack dense 3D geometry and thermal signals, which are believed to be critical for a deeper understanding of affective states. Furthermore, agreement-oriented datasets like HMI-Mimicry  \cite{bilakhia2015mahnob} rely solely on audio and 2D video, omitting the richer visual and physiological channels which are believed to underpin the complex expression of stance. In short, the lack of high-fidelity multimodal datasets that capture the full spectrum of dyadic communication impedes research in the field.  

To address this critical gap, we create and present a new, large-scale multimodal dataset of face-to-face dyadic interactions. Our data collection paradigm was designed and implemented by bilateral captures of conversational dyads, with a set of well-synchronized sensors, including high-fidelity visual sensing (high-resolution 2D, dense 3D, and thermal imaging), vocal audio sensing, and affective interaction sensing (a set of physiological signals, e.g., {\em Photoplethysmogram (PPG), Elecreodermal Activity (EDA), Heart-rate (HR), Blood Pressure (BP), and Respiration-rate (RR)}).
%
By deploying a state-of-the-art, multi-sensor suite within an uniquely designed dyadic context, 
we have developed three dyadic interaction tasks and captured data from 45 interacting dyads (90 subjects) to support dynamic modeling of interpersonal influence across expressive behavior and physiology, as well as analysis of emotional synchrony.  
Figure~\ref{fig:dyadic_intro} shows the multimodal bilateral data collection at work in the scenario of dyadic conversation with a pair of participants.
Our dataset provides a unified and comprehensive resource to study how external, observable behaviors align with internal, physiological states during moments of agreement, disagreement, and neutrality.

Our work makes the following primary contributions:
\begin{enumerate}
\item We develop and share a new large-scale, multimodal dataset of dyadic conversations explicitly designed to elicit naturalistic vocal/visual/physiological behaviors associated with agreement, disagreement, and neutrality, resulting in a total of 270 sets of multimodal sequences of 45 dyads (i.e., 90 subjects), with 1400+ minutes (i.e., 2.1 millions of frames) of dyadic interactive communications.  
\item The dataset features an unprecedented combination of synchronously recorded data streams for both participants: high-resolution 2D videos, high-fidelity dynamic 3D facial geometry, and high-resolution thermal videos and temperature data, complemented by synchronized audio, texts, and physiological signals, including PPG, heart-rate, blood-pressure, EDA, and respiration.

\item An unique IRB-approved protocol is designed and implemented for eliciting stance-related behaviors of dyads effectively,  which is essential for capturing authentic and unscripted interactions of dyads, advancing the multimodal modeling of social interaction never before.
\end{enumerate}
\section{Related Work}
\label{sec:related}

\begin{table*}[ht]
\centering
\caption{Comparative Analysis of Publicly Available Multimodal Dyadic Datasets.}
\label{tab:dataset_comparison}
\resizebox{\textwidth}{!}{%
\begin{tabular}{P{2cm}|P{1.2cm}|P{2.3cm}|P{3.5cm}|P{2.5cm}|P{3.2cm}|P{2.5cm}}
\hline
\textbf{Dataset Name} & \textbf{Dyads} & \textbf{Participants} & \textbf{Modalities} & \textbf{Interaction Task} & \textbf{Key Annotations} & \textbf{Spontaneous/Acted} \\
\hline
IEMOCAP~\cite{busso2008iemocap} & 5 & 10 Actors & Audio, Video, Motion Capture (Face, Hands) & Scripted/Improvised Scenes & Emotion Categories, Valence, Arousal & Acted \\
\hline
RECOLA~\cite{ringeval2013introducing} & 23 & 46 Students & Audio, Video, ECG, EDA & Remote Collaborative Task & Valence, Arousal & Spontaneous \\
\hline
SEMAINE~\cite{mckeown2011semaine} & $\sim$150 & Users \& Operator & Audio, Video & Human-Agent Conversation & Emotion Dimensions, FACS & Spontaneous  \\
\hline
MMDB~\cite{rehg2013decoding} & 121 & Adult-Child Dyads & Audio, Video (multi-view), EDA, Accelerometry & Semi-structured Play & Engagement, Behaviors (gaze, smile) & Spontaneous \\
\hline
HMI-Mimicry~\cite{bilakhia2015mahnob} & 54 & 12 confederates \& 48 counterparts & Audio, Video (multi-view) & Political Discussion / Tenancy Negotiation & Mimicry, Dialogue Acts, Head/Hand Gestures & Spontaneous \\
\hline
\textbf{Inter-Stance} & \textbf{45} & \textbf{90} & \textbf{2D/3D Video, Thermal, HR, BP, PPG, EDA, Audio, Text} & \makecell{\textbf{Face-to-Face}\\\textbf{Discussion}} & \textbf{Stance (Agree, Disagree, Neutral), Synchrony} & \textbf{Spontaneous} \\
\hline
\end{tabular}%
}
\vspace{-3mm}
\end{table*}

There are a number of datasets existed for study of emotion in dyads and interpersonal stances. 

\subsection{Datasets for Emotion and Affect in Dyads}
Several foundational datasets have been instrumental in the study of emotion. 
The IEMOCAP~\cite{busso2008iemocap} database is a seminal resource containing data from actors in dyadic sessions, with modalities including audio, video, and motion capture of the face and hands.  
While its motion capture provides valuable data on gestures, it lacks the dense 3D facial geometry needed for fine-grained expression analysis and does not include thermal imaging.  
Similarly, the MSP-IMPROV~\cite{busso2016msp} corpus features dyadic improvisations but is limited to audio and video, omitting the advanced visual sensors required for deeper analysis.  
The MELD~\cite{poria2018meld} dataset uses dialogues from a TV series, offering multi-party audio and video, but the interactions are scripted and lack specialized sensor data.

Other datasets focus on interactions within a specific context. 
The RECOLA~\cite{ringeval2013introducing} database is a key resource for remote collaboration, including video and physiological signals (ECG, EDA). 
However, the remote video-conference setting fundamentally alters non-verbal cues compared to face-to-face interaction, and it lacks 3D and thermal data.  
The SEMAINE~\cite{mckeown2011semaine} database focuses on human-agent, not human-human, interactions, which involve different social dynamics.  
The MMDB~\cite{rehg2013decoding} dataset is a unique resource for studying adult-child interactions, including multi-view video, but its focus on developmental psychology makes it unsuitable for analyzing adult conversational stance.  
BP4D series databases~\cite{bp4d2014,bp4d+_2016,bp4d++2023} contain conversational interactions where an interviewer engages a participant to elicit a range of spontaneous emotions.  In these recordings, the comprehensive sensor suite is focused on capturing data solely from the participant; the interviewer is part of the elicitation protocol but is not recorded.

\subsection{Datasets for Interpersonal Stance and Mimicry}

Research in computational stance detection has predominantly focused on textual data, leading to the development of several key benchmark datasets.
SemEval-2016 Task 6~\cite{mohammad2016semeval}, and P-Stance~\cite{li2021p} provided corpus of tweets for target-based stance detection.
A recent Mutimodal dataset~\cite{liang2024multi} have advanced the field by providing a large-scale, multimodal (text and image) corpus from social media for complex multi-target stance detection.
In contrast, our work provides synchronous, dyadic, face-to-face conversations, enabling a direct analysis of the real-time dynamics inherent in live interactions.

The MHi-Mimicry~\cite{bilakhia2015mahnob} database was explicitly designed to analyze behavioral mimicry during discussions involving agreement and disagreement, using 18 synchronized audio and video sensors.  
Despite its highly relevant design, its critical limitation from a vision perspective is the complete absence of 3D, thermal, and physiology data.
Furthermore, a significant methodological concern is the participant composition. 
Of the 60 subjects, 12 were confederates who participated in multiple sessions. 
This repetition may have constrained the spontaneity of the interactions, potentially reducing the authenticity of the free-form speech captured.

\subsection{Positioning Our Dataset}

We introduce Inter-Stance, a novel corpus for investigating the embodied, interactive nature of stance in face-to-face conversation. To capture the dynamics of agreement and disagreement, we adapt a state-of-the-art visual and physiological sensing suite to simultaneously record both members of a dyad, a methodology typically reserved for single-participant analysis. 
As summarized in Table~\ref{tab:dataset_comparison}, Inter-Stance fills a crucial research gap by providing the first dataset of its kind, enabling quantitative analysis of behavioral and autonomic correlates of interpersonal stance.


\section{Data Acquisition}
\label{sec:data}

The creation of this dataset was guided by two principles: methodological rigor in eliciting naturalistic behavior and technical precision in capturing high-fidelity, synchronized multimodal data with a set of advanced visual and physiological sensors.

\begin{table}[ht]
\centering
\caption{Ethnic Distribution of Participants}
\begin{tabular}{l c c}
\toprule
\textbf{Ethnic/Racial Ancestry} & \textbf{Number} & \textbf{Proportion (\%)} \\
\midrule
White & 41 & 45.6 \\
East Asian & 26 & 28.9 \\
Latino/Hispanic & 10 & 11.1 \\
Black & 9 & 10.0 \\
South Asian & 4 & 4.4\\
\midrule
\textbf{Total} & \textbf{90} & \textbf{100.0} \\
\bottomrule
\end{tabular}
\label{tab:ethnic_distribution}
\vspace{-3mm}
\end{table}

\subsection{Participant}

Participants have been recruited from a university campus and the local community. 
The target sample size is 45 dyads (90 participants), providing sufficient data for training robust vision models. 
 
Demographic information, including age, gender, and ethnic/racial ancestry, was collected to allow for subgroup analyses. 
Table~\ref{tab:ethnic_distribution} shows the ethnic distribution.
All procedures have been approved by the university's Institutional Review Board (IRB), and all participants have provided written informed consent.

\subsection{Experimental Protocol}

A novel experimental protocol was designed to reliably elicit multimodal behaviors associated with agreement, disagreement, and neutrality without sacrificing spontaneity.

{\bf Overall Procedure}: Each dyad will participate in a single 45-minute session consisting of three 5-minute discussion blocks, each targeting one stance: Neutral, Disagree, or Agree. Procedure:
 After the participants give informed consent, the experimenter will provide instructions and put the dyads at ease. 
 The participants will participate in three tasks designed through our Stance Elicitation Mechanism as described in follows. 
 A pre-survey is conducted by the two participants given a list of possible topics, topics will be scored to measure the intensity of their feelings about each.  
 The experimenter then selects neutral topic, topic of strong disagreement, and topic of mutual enjoyment or agreement. The topic discussions each last approximately five minutes.

{\bf Stance Elicitation Mechanism:} The key to eliciting genuine stance is the selection of discussion topics based on participants' pre-existing opinions.

\begin{enumerate}

\item Pre-session Survey: Before arriving for the study, each participant completes an online survey. 
This survey contains a series of statements on various topics, and participants rate their opinion on a 13-point scale ranging from -6 (strongly disagree) to +6 (strongly agree), with 0 indicating a neutral stance. 
The topics consist of Sports, Religion, Relationships, Education, Media, Alcohol/Drugs, Technology, Health, Money, Science, Politics, and Family.
\item Topic Assignment: Upon a dyad's arrival, an experimenter analyzes their survey responses to strategically select three discussion topics. 
A topic where the participants expressed opposing views was chosen for the Disagree condition. 
A topic where they expressed similar views is selected for the Agree condition. 
A non-controversial, factual topic is assigned for the Neutral condition.
\item Instructions: Participants will be instructed simply to ``Please discuss the following topic." 
They are not informed of their partner's stance, preserving the natural discovery process. 
This design is crucial for capturing genuine interactive behaviors.

\end{enumerate}
Table~\ref{tab:topic_examples} provides examples of the topics used to guide the discussions.
Full topic list will be attached in supplementary material.

\begin{table}[t]
\centering
\caption{Example Topic Assignments Based on Pre-Session Survey Scores}
\label{tab:topic_examples}
\resizebox{\columnwidth}{!}{%
\begin{tabular}{P{3.5cm}|P{1.6cm}|P{1.6cm}|P{1.2cm}}
\hline
\textbf{Topic} & \textbf{Subject 1} & \textbf{Subject 2} & \textbf{Stance} \\
\hline
"Social media platforms do more harm than good for society." & Agree (+6) & Disagree (-5) & Disagree \\
\hline
"Smoking should be banned on all college campuses." & Agree (+5) & Agree (+6) & Agree \\
\hline
"Streaming music services are negatively impacting the music industry." & Disagree (-4) & Agree (+4) &  Disgree \\
\hline
"Cities should invest more in public parks and green spaces." & Agree (+4) & Agree (+4) &  Agree \\
\hline
"Depression is caused by a chemical imbalance in the brain" & Neutral (0) & Neutral (0) & Neutral \\
\hline
\end{tabular}%
}
\vspace{-3mm}
\end{table}

\subsection{Multimodal Recording System and Synchronization}

This data acquisition system is designed to simultaneously and synchronously record a complete set of visual, auditory, and physiological signals from two participants. 
For each participant, the setup consists of a 3D dynamic (a.k.a. 4D) imaging system, a thermal sensor, an audio sensing device, and a physiological sensing system, all positioned to ensure high-quality data acquisition during face-to-face interaction.
The thermal camera, mounted on a tripod alongside the 3D imaging system, maintains a fixed distance from the subject.

System synchronization is critical for analyzing interpersonal dynamics across modalities. 
As each sensor is controlled by a dedicated machine, a custom program was developed to manage the recording process. 
This is achieved through a master machine that sends a trigger signal concurrently to the 3D imaging, thermal, audio collection device, and physiological systems, ensuring all data streams are aligned with microsecond-level precision from start to finish. 

\begin{figure}[th]
  \centering
  \begin{subfigure}[t]{0.99\linewidth}
    \includegraphics[width=\textwidth]{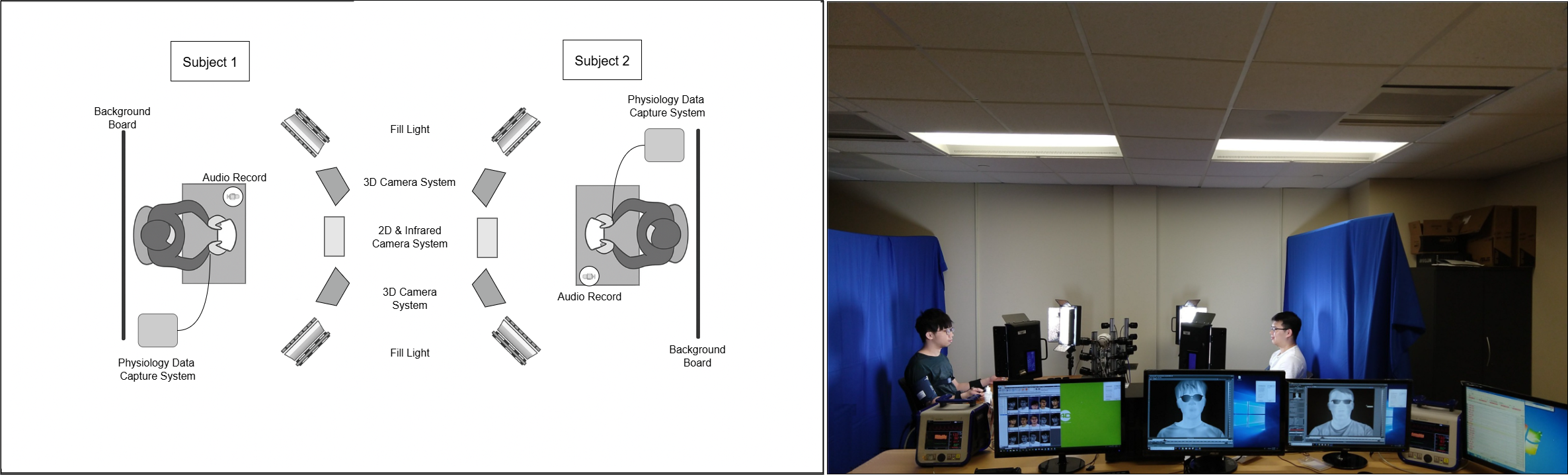}
  \end{subfigure}
  \caption{Dyadic data collection system at work. Note that the bilateral visual sensors (4D imaging systems) do not block the sights of both dyads.}
  \label{fig:tsne}
  \vspace{-3mm}
\end{figure}

\subsubsection{4D Dynamic Imaging System}

High-resolution 3D facial geometry and the corresponding 2D texture video are captured by a dual DI4D stereo imaging system operating at 25 fps \cite{Di4D}.
Each reconstructed 3D model consists of approximately 30,000 to 50,000 vertices, achieving a root mean square (RMS) accuracy of 0.2 mm, which allows for the detailed analysis of fine-grained geometric changes in facial musculature.
The integrated 2D color camera captures texture video at a resolution of $1536 \times 2048$ pixels, providing essential appearance-based data that complements the 3D geometry.
The 25 fps frame rate for both 3D and 2D streams is aligned with the thermal sensor's capture rate.

\subsubsection{Audio System} 
To capture high-quality, channel-separated audio, recording is handled independently of the DI4D imaging system. 
Each participant is equipped with an individual Yeti stereo microphone. 
Audio is captured at a sampling rate of 48 kHz with 16-bit depth. 

\subsubsection{Thermal Sensor}

A FLIR A655sc long-wave infrared camera \cite{FLIR} captures facial skin temperature dynamics at a resolution of $640 \times 480$ pixels and a frame rate of 25 fps.
The sensor operates within a temperature range of -40$^{\circ}$C to 150$^{\circ}$C and a spectral range of $7.5-14.0\mu m$.
This modality provides a unique, non-contact visual channel into autonomic nervous system activity, as changes in stress and arousal manifest as shifts in facial blood flow patterns that are invisible to standard cameras but detectable in the thermal spectrum.

\subsubsection{Physiological Sensing System}

A Biopac MP150 data acquisition system \cite{Biopac} is used to record a suite of physiological signals from each participant at a sampling rate of 1000 Hz.
The recorded signals include:

\begin{itemize}
    \item \textbf{Cardiovascular Signals:} Blood Pressure (BP) is captured using a Biopac NIBP100D monitoring system, which includes a finger unit placed on the index and middle fingers and an inflatable cuff on the arm for calibration. 
    The system's measurement capacity for blood pressure is [-25 mmHg, 300 mmHg]. 
    From the continuous BP waveform, several key metrics are derived, including systolic BP (mmHg), diastolic BP (mmHg), and pulse rate (beats/minute). 
    The photoplethysmogram (PPG) signal (V) is also captured from another index finger sensor on the other hand. Heart rate (HR) is derived with a capacity of [30, 300 beats/minute].
    
    \item \textbf{Respiration:} A respiration belt (Biopac RSP100C) worn around the participant's chest measures thoracic expansion and contraction as a voltage signal. From the peak count of this signal, the respiration rate (breaths/minute) is derived, with a measurement capacity of [0, 200 breaths/minute].
    
    \item \textbf{Electrodermal Activity (EDA):} EDA, a direct indicator of sympathetic arousal, is measured in microSiemens (${\mu}S$). The signal is captured via two leads placed on the participant's right palm, connected to a wrist unit.
\end{itemize}

\subsection{Self-Report}

A post-experiment self-report questionnaire was administered immediately after all the conclusion of the dyadic interactions. 
The survey prompted each participant to retrospectively evaluate their feelings for every conversations individually. 
For a predefined set of emotions, including Engaged, Pleased, Excited, Offened, Indifferent, Bored, and Embarrassed, participants rated the peak intensity they felt on a 4-point scale: None, Mild, Moderate, or Intense. 

\section{Data Annotation}
To maximize the utility of the dataset for the research community, we have generated a rich set of metadata and annotations. 
This includes semi-manual annotations of expression, as well as automatically derived features such as facial landmarks and the speech text from audio.
\\\\
\noindent\textbf{Stance:}
The stance labels (Agreement, Disagreement, Neutral Stance) for this corpus were assigned at the conversation level, based on the experimental design.
Prior to the interaction, participants completed a survey to determine their opinions on a set of topics. 
Dyads were then formed and assigned a topic with the explicit goal of eliciting one of the three stances. 
This method provides a coarse, session-wide ground-truth label for each dyadic interaction, treating the entire conversation as a single data point for a given stance.
\vspace{1mm}

\noindent\textbf{Expression:}
Facial expressions are labeled at the frame level, where we applied three FER models~\cite{serengil2024lightface,savchenko2022classifying,chang2023libreface} to extract their logits.
To make the label more robust, we sum these three logits and predict the label using a softmax activation function.
Additionally, three professional coders manually reviewed samples of frames and corrected errors found within them.
\vspace{1mm}

\begin{figure}
    \centering
    \begin{subfigure}[t]{\linewidth}
    \includegraphics[width=0.97\linewidth]{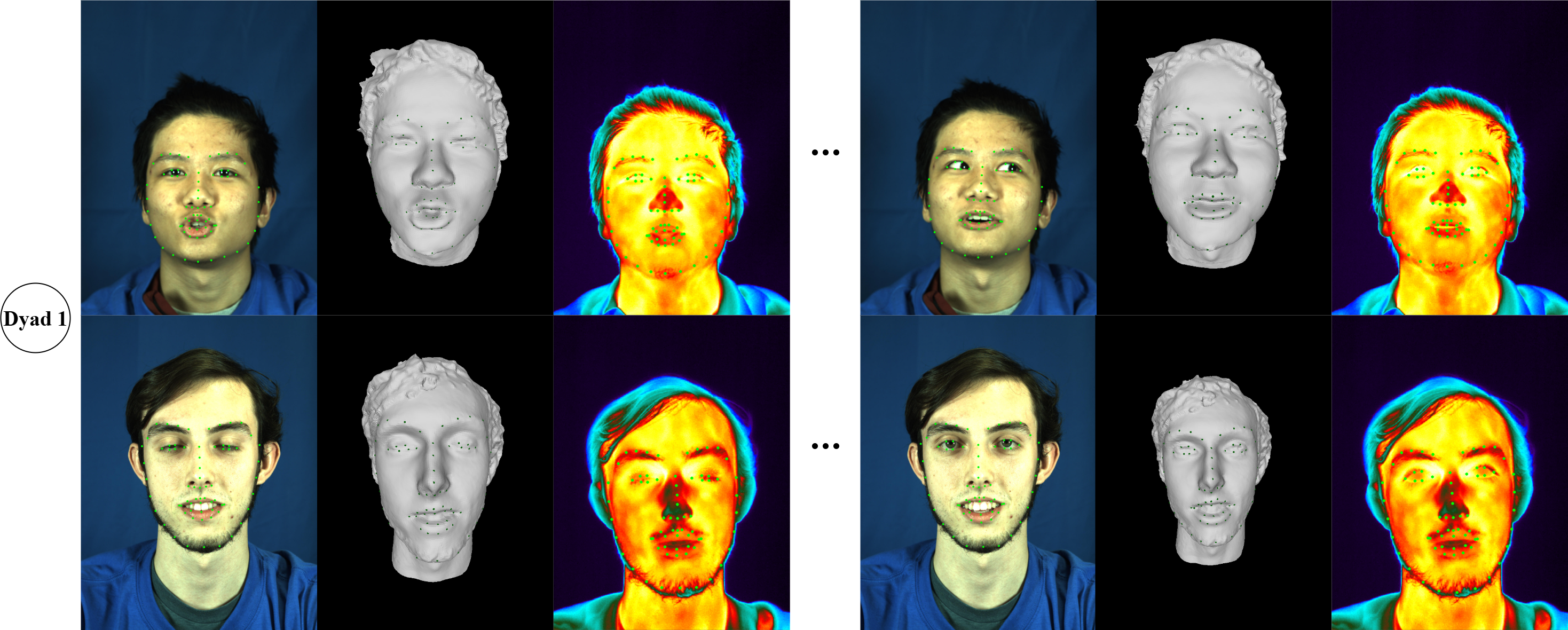} 
    \caption{Acquainted Dyads}
  \end{subfigure}
  \begin{subfigure}[t]{\linewidth}
    \includegraphics[width=0.97\linewidth]{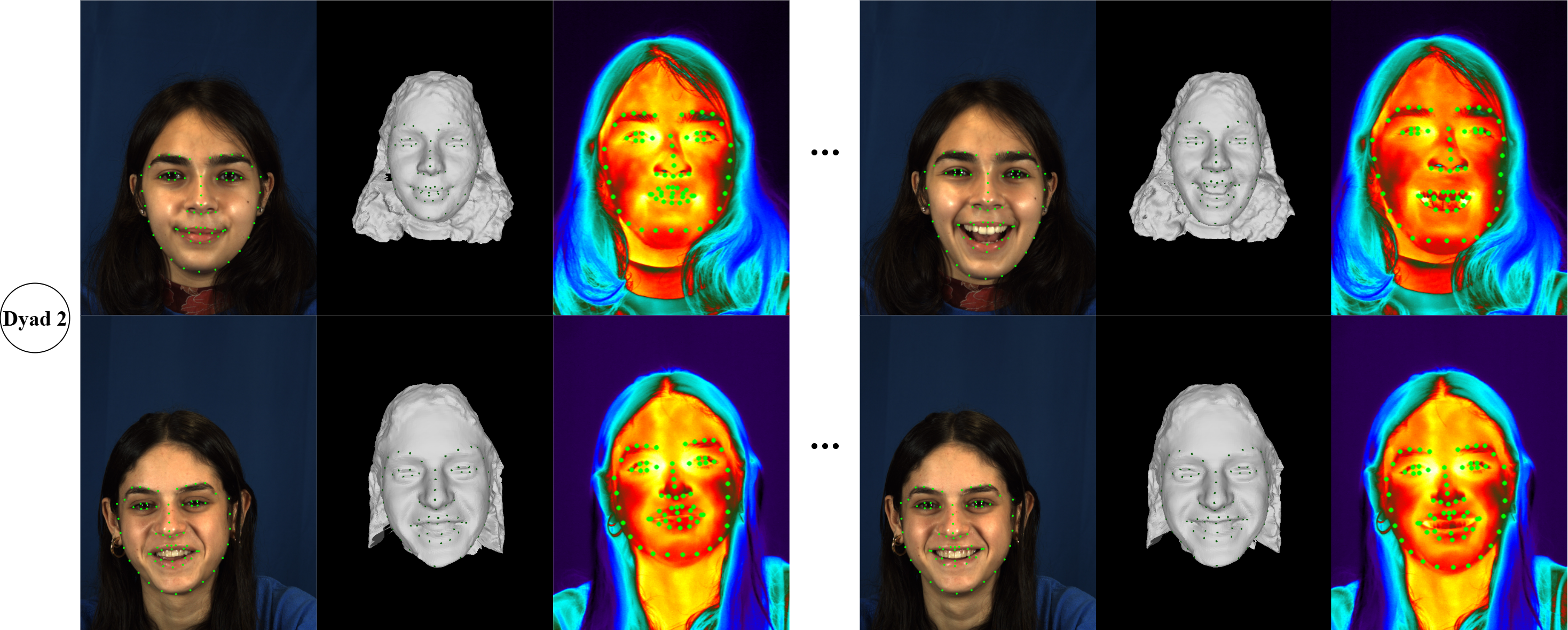} 
    \caption{Unfamiliar Dyads}
  \end{subfigure}
    
    \caption{Dyads Sequence Samples of Multimodal Feature Points}
    \label{fig:landmarks}
    \vspace{-3mm}
\end{figure}

\noindent\textbf{Feature Point Tracking:}
To facilitate computational analysis, facial feature points (landmarks) were automatically tracked across multiple visual modalities for each frame of the recordings. 
Fig.~\ref{fig:landmarks} shows several dyads' sample frames with
 tracked points.
\subsubsection{3D Feature Tracking}
We tracked 68 facial landmarks directly on the 3D geometric mesh sequences. 
This was achieved using a shape index-based statistical shape model (SI-SSM)~\cite{canavan2015landmark}, which is robust to a wide range of dynamic expressions and head movements.  
The model takes advantage of global facial structure and local patch information to accurately track key points around the eyes, nose, mouth, and eyebrows. 

\subsubsection{2D Feature Tracking}
For the corresponding 2D texture videos, 68 facial landmarks were tracked using an automated method based on ~\cite{PFL}. 
The backbone is Res2Net50~\cite{gao2019res2net}.

\subsubsection{Thermal Feature Tracking}
64 corresponding facial landmarks were also tracked on the thermal video streams.  
The thermal data was first pre-processed to enhance local contrast. Subsequently, a Constrained Local Model (CLM) was applied to detect and track the landmarks, using the mean face shape for initialization and refining the fit with an Active Shape Model~\cite{cootes1995active}. 
\vspace{1mm}


\noindent\textbf{Audio to Text:}
The high-quality, channel-separated audio recorded from each participant's microphone was processed to generate time-aligned transcripts. 
An automatic speech recognition (ASR) system~\cite{radford2023robust} was employed to convert the spoken dialogue into text. 
This process results in a separate, synchronized transcript for each participant, allowing researchers to analyze the lexical content of the conversation in conjunction with the rich visual and physiological data streams.

\section{Experiments and Validation}
To validate the usefulness of the data, we conduct a series of experiment on it, include stance recognition, face expression recognition, and statical analysis on various modalities.
Table~\ref{tab:results_summary} shows Stance Recognition and FER results across modalities.

\subsection{Stance Recognition on Physiological Data}

This experiment aimed to determine the efficacy of physiological signals as indicators of conversational stance. The input data consisted of 20 channels of physiological signals recorded concurrently from both subjects in the dyad. The signals included respiratory, photoplethysmography (PPG), blood pressure (BP), electrodermal activity (EDA), and their derivatives, such as respiratory rate, pulse rate, and diastolic/systolic BP. 
The raw data underwent a standardized preprocessing pipeline: 1) Downsampling: signals were downsampled from 1000Hz to 200Hz to reduce computational load while preserving relevant features.
2) Imputation: missing values and zero-crossings were replaced with the mean value of the respective channel.
3) Standardization: a Z-score normalization was applied to each channel independently to scale the data.
4) Windowing: the continuous signals were segmented into 1-second non-overlapping windows, which served as the input samples for our models.

We systematically evaluated four deep learning architectures of increasing complexity: RNN~\cite{sherstinsky2020fundamentals}, LSTM~\cite{sherstinsky2020fundamentals}, Attention LSTM~\cite{wang2016attention}, and Transformer~\cite{vaswani2017attention}. We employed two distinct evaluation protocols: 
1) an Intra-subject scheme, where models were trained and evaluated on each dyad's data separately using 5-fold cross-validation. 
2) an Inter-subject scheme, where a single, generalized model was trained on the data from all dyads combined.

\begin{figure}
    \centering
    \includegraphics[width=0.95\linewidth]{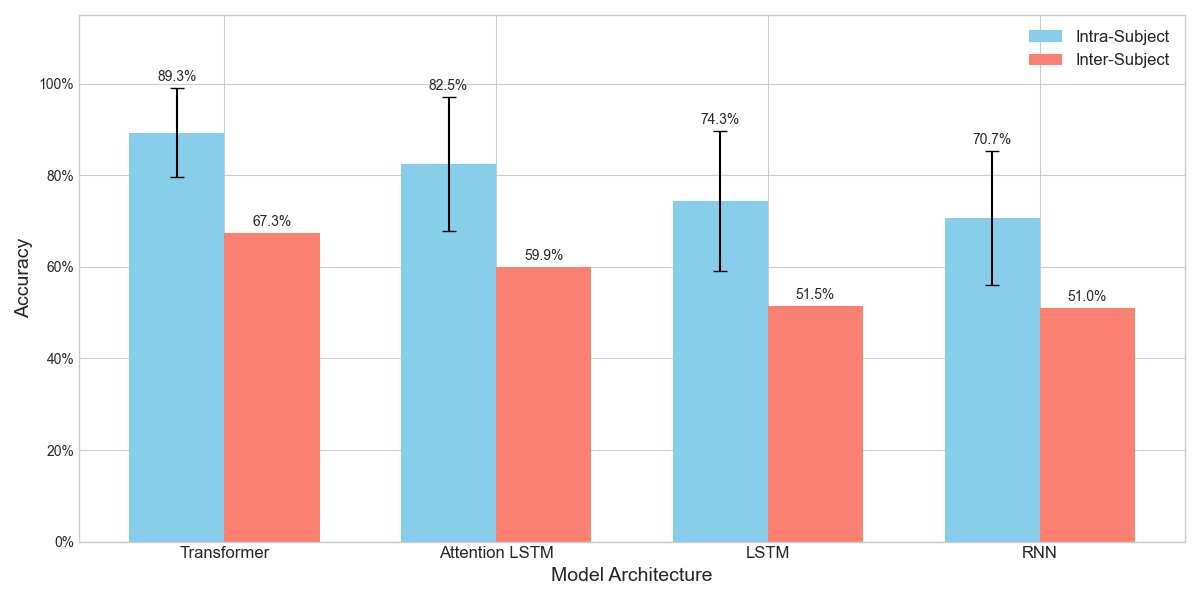}
    \caption{Stance Recognition Accuracy from Physiological Signals}
    \label{fig:physiology_result}
    \vspace{-3mm}
\end{figure}

Our analysis is illustrated in Fig.~\ref{fig:physiology_result}, reveals two primary findings. 
First, model performance scales with architectural complexity, with Transformer-based models consistently achieving the highest accuracy. 
Second, and more critically, we found that physiological expressions of stance are highly subject-dependent. 
This is evidenced by both a substantial performance gap (22\% for the Transformer) between personalized intra-subject models and the single, generalized inter-subject model, as well as high performance variance across the different dyads. 
These results confirm that while generalizable physiological markers for stance exist, the most potent and reliable signals are highly individualized, highlighting inter-subject variability as a central challenge in physiological computing.

\subsection{Stance Recognition on Video}

To complement our physiological analyses, we investigated stance recognition from video, hypothesizing that visual expressions of stance are highly subject-dependent, similar to autonomic responses. 
Our initial experiment, using a subject-independent split to test generalization to unseen individuals, yielded a modest accuracy of 48.0\%, suggesting that individual differences were a key factor.

To directly test this hypothesis in the visual domain, we conducted a second experiment following the inter-subjects protocol in the physiology experiment, where the model was exposed to all subjects during training. 
This model achieved a dramatically higher accuracy of 85.6\%. 
This substantial increase of nearly 40 percentage points from the subject-independent result provides powerful, direct evidence that visual expressions of stance are highly personalized. 
Furthermore, the 85.6\% accuracy significantly surpasses the 67.3\% achieved by the equivalent general physiological model. 
This suggests that when person-specific cues can be learned, the rich and direct communicative signals available in facial video (e.g., head movements, expressions) are substantially more predictive of stance than the more indirect autonomic signals from physiology. 

Moreover, a multimodal framework, concatenating the visual and physical features, is designed for stance recognition, achieving 89.8\% accuracy, proving the advantage of using two modalities.
The model architecture and settings can be seen in Supplementary Materials.
More modalities will be included in future work.


\begin{table}[t!]
\centering
\caption{Summary of Stance Recognition and FER Performance Across Modalities and Methods. 3 setups: Inter-Subjects; Intra-Subjects; Cross-Subjects.}
\begin{tabular}{lllr}
\toprule
\textbf{Task} & \textbf{Modality} & \textbf{Method / Setup} & \textbf{Accuracy (\%)} \\
\midrule
Stance  & Text & BERT (Inter) & 63.5 \\
 & Physiology & Transformer (Inter) & 67.3 \\
 & Video & CLEF (Inter)& 85.6 \\
 & Video + Physiology & LateFusion (Inter) & \textbf{89.8} \\
\cmidrule{2-4}
& Physiology & Transformer (Intra)& 89.3 \\
\cmidrule{2-4}
& Video & CLEF (Cross)& 48.0 \\
\hline
FER & Video & CLEF (Cross)  & 73.5 \\
\bottomrule
\end{tabular}
\label{tab:results_summary}
\vspace{-3mm}
\end{table}

\subsection{Facial Expression Analysis}

For each stance category (Neutral Stance, Disagreement, Agreement), we collected the distribution of seven emotion classes (Neutral, Happiness, Sadness, Anger, Disgust, Surprise, Fear). 
To ensure comparability across stance categories, raw counts were normalized into percentage distributions. 
We then visualized the results using both grouped and stacked bar charts, where each stance category is represented as a 100\% stacked distribution of emotions. This design enables us to highlight subtle changes in emotional composition across stance types.

\begin{figure}[t]
  \centering
  \begin{subfigure}{0.59\linewidth}
    \includegraphics[width=\textwidth]{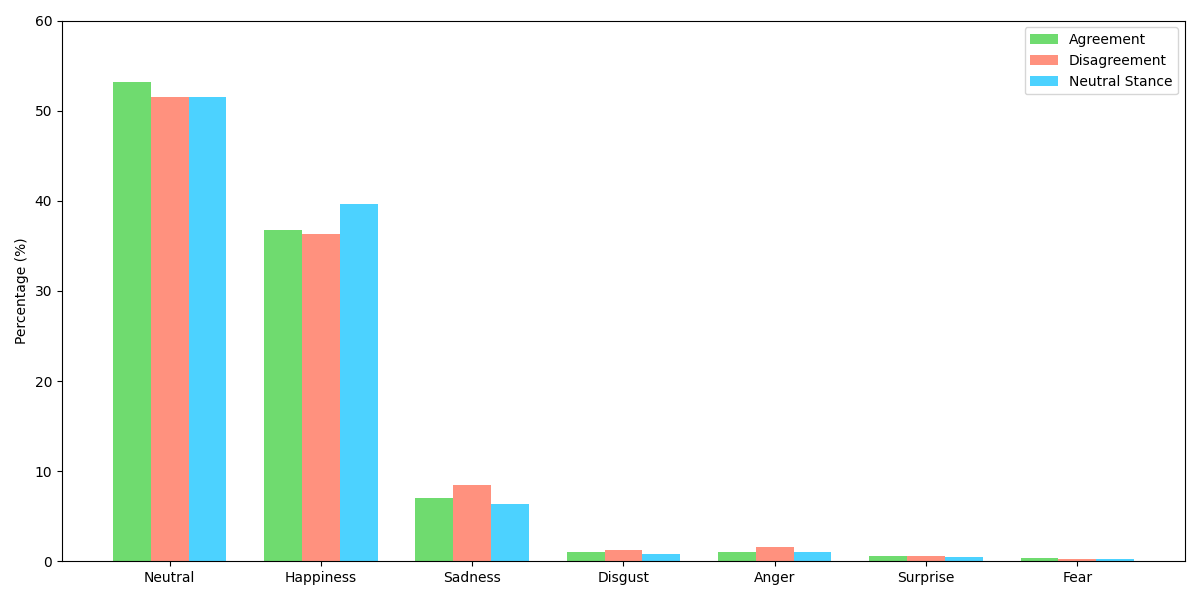}
    \caption{Comparison of emotion distributions across stance categories}
  \label{fig:fer_distribution_a}
  \end{subfigure}
  \hfill
  \begin{subfigure}{0.39\linewidth}
    \includegraphics[width=\textwidth]{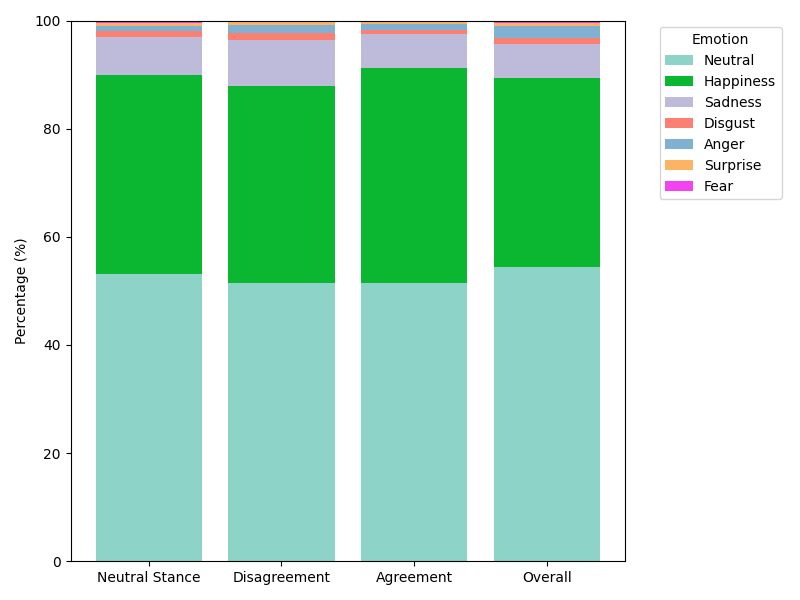}
    \caption{Emotions within each stance category}
    \label{fig:fer_distribution_b}
  \end{subfigure}
  \caption{Visualization of emotion distribution across three stance categories (Agreement / Disagreement / Neutral Stance)}
  \label{fig:fer_distribution}
  \vspace{-3mm}
\end{figure}

Fig.~\ref{fig:fer_distribution_a} compares emotion distributions across three stance categories (agree, disagree, neutral), and Fig.~\ref {fig:fer_distribution_b} shows stacked percentage bars in each stance category. 
Clear differences are observed:
1) \B{Agreement} contains a higher proportion of happiness, while negative emotions remain relatively rare.
2) \B{Disagreement} is enriched in negative emotions such as Sadness, Anger, and Disgust, significantly higher than in the other stances.
3) \B{Neutral stance} is dominated by Neutral emotion with a notable presence of Happiness, reflecting an overall more balanced and flat affective profile.

Overall, the pattern is intuitive and consistent: agreement tends to be happier, disagreement more negative, and neutrality more emotionally flat. 
This validates that emotion signals align with stance polarity and can provide complementary cues for stance recognition.

\noindent\B{Facial Expression Recognition}
We further conducted experiments on facial expression recognition (FER) to examine the relationship between stance and affective signals. 
Since the original dataset exhibited severe class imbalance, with Neutral and Happiness dominating the distribution and negative emotions sparsely represented, we grouped emotions into three categories: Happiness, Neutral, and Negative (including Anger, Sadness, Disgust, Fear, Surprise) and sample them into balance. 
We use CLEF~\cite{zhang2023weakly} as the baseline model, with a subject-independent setting, achieving 73.5\% accuracy.
CLEF is pretrained on BP4D+, with a weakly-supervised setting.
This result demonstrates that, the baseline model captures meaningful affective information. 
Nonetheless, the moderate performance indicates that distinguishing subtle negative expressions remains challenging, and more advanced models or temporal context exploitation may be necessary to further improve recognition accuracy.

\subsection{Analysis of Facial Mimicry}

\begin{figure}[b]
     \centering
     \begin{subfigure}[b]{0.49\linewidth}
         \includegraphics[width=\textwidth]{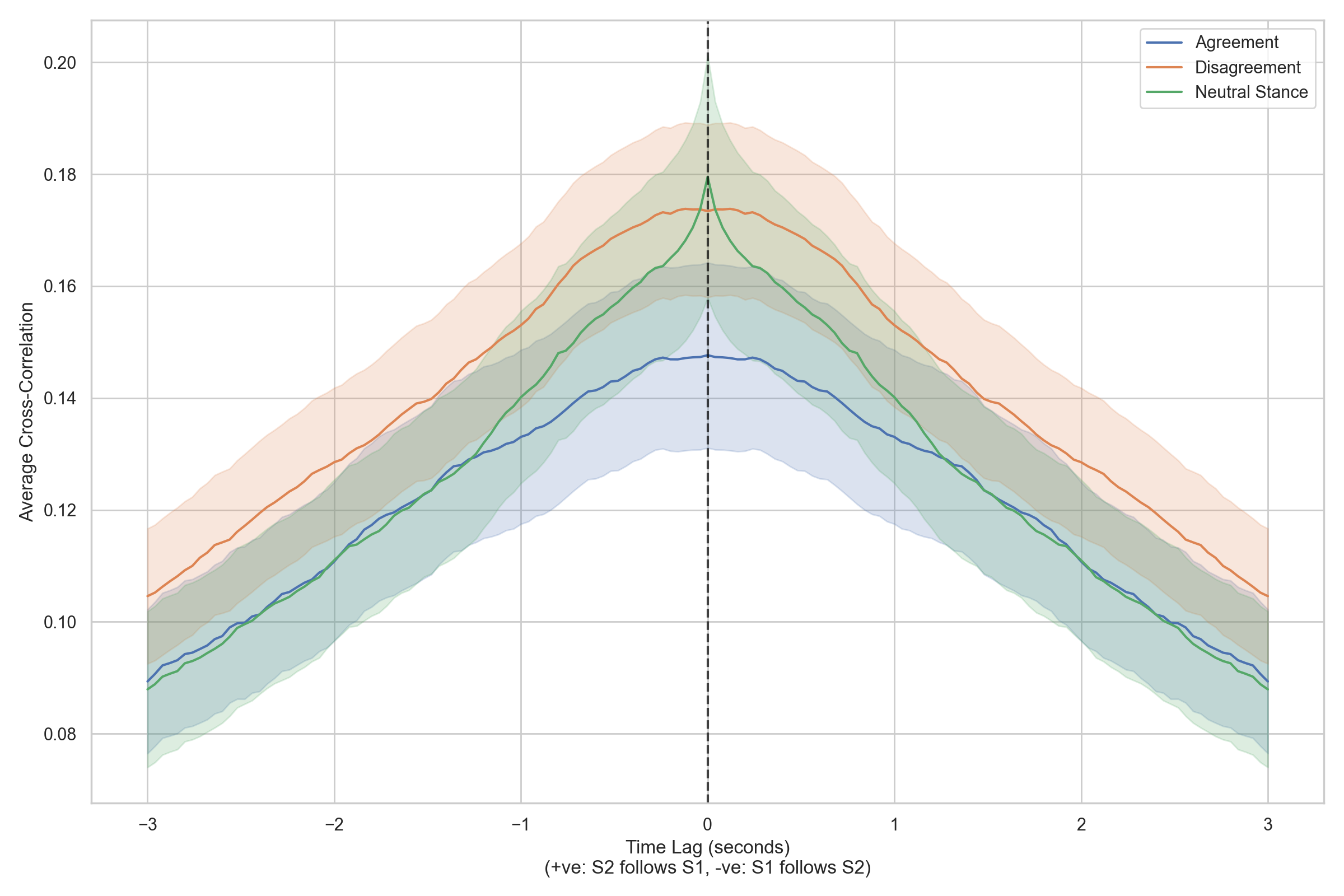}
         \caption{Happiness}
     \end{subfigure}
     \begin{subfigure}[b]{0.49\linewidth}
         \includegraphics[width=\textwidth]{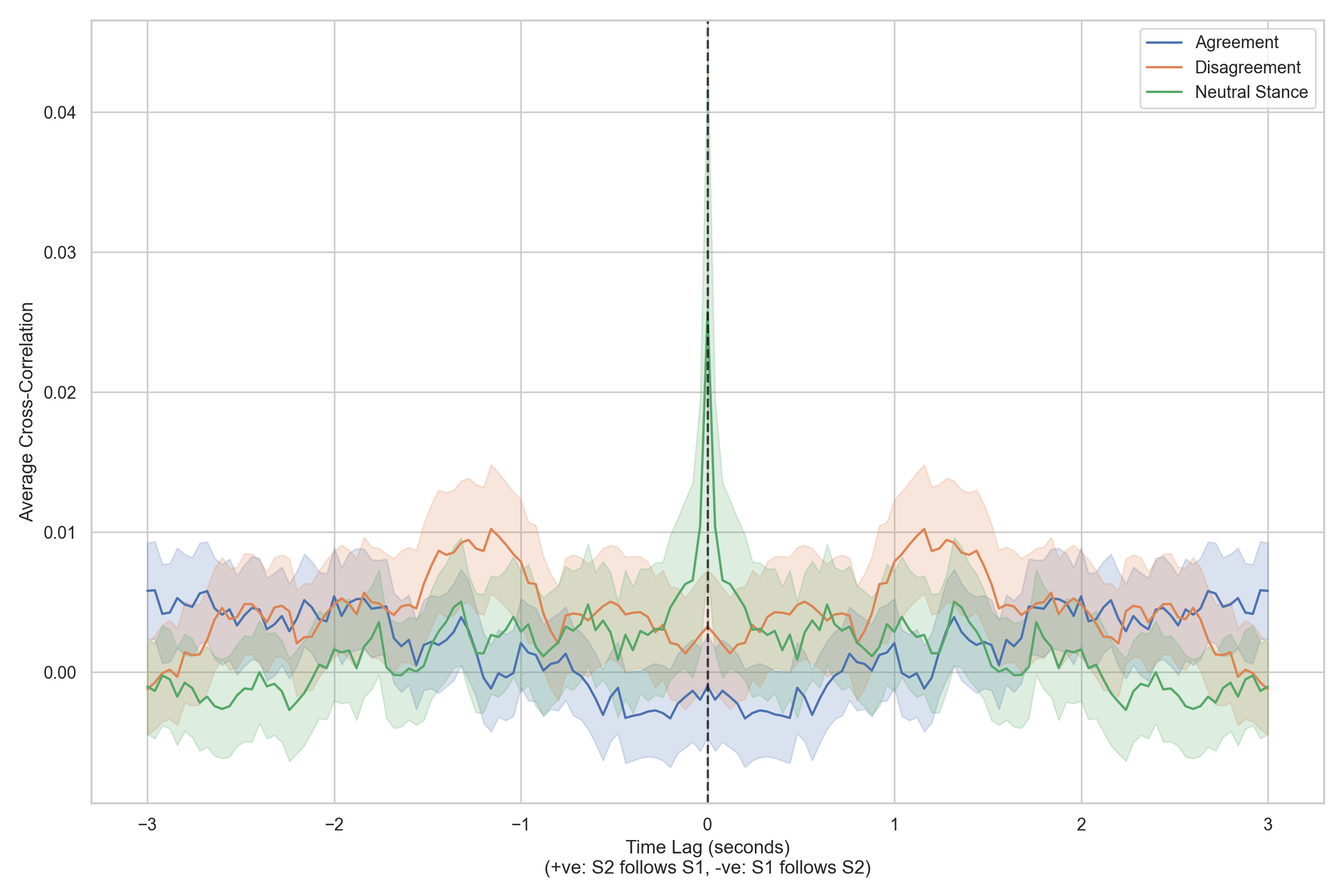}
         \caption{Negativeness}
     \end{subfigure}
    \caption{Facial Mimicry of Emotion by Conversational Stance}
    \label{fig:mimic_stance}
\end{figure}
To investigate facial mimicry, we performed a lagged cross-correlation analysis on frame-level emotion predictions for Happiness and a composite Negative emotion category. The results, visualized in Fig.~\ref{fig:mimic_stance}, reveal two functionally distinct mimicry patterns. 
We observed a strong affiliative mimicry for Happiness, evidenced by a significant correlation peak (r $\approx$ 0.03) at a time lag of approximately +0.5 to +1.0 seconds. 
The strength of this mirroring effect was heavily modulated by the conversational context, being strongest in Agree stances, demonstrating its role in building social rapport. In stark contrast, we found a weaker, confrontational mimicry for Negative emotions that was exclusively present within disagreeable conversations (r $\approx$ 0.02) at a similar time lag. 
This suggests that while smile mimicry is a broad, pro-social behavior used to reinforce consensus, the mimicry of negative affect is a rarer, context-specific signal tied to conflict. 
The clear dichotomy between these patterns underscores that facial mimicry is not a simple reflex, but a sophisticated social signaling mechanism deployed differently based on emotional valence and conversational context.

\subsection{Audio Feature Analysis}

\begin{figure}
    \centering
    \includegraphics[width=0.95\linewidth]{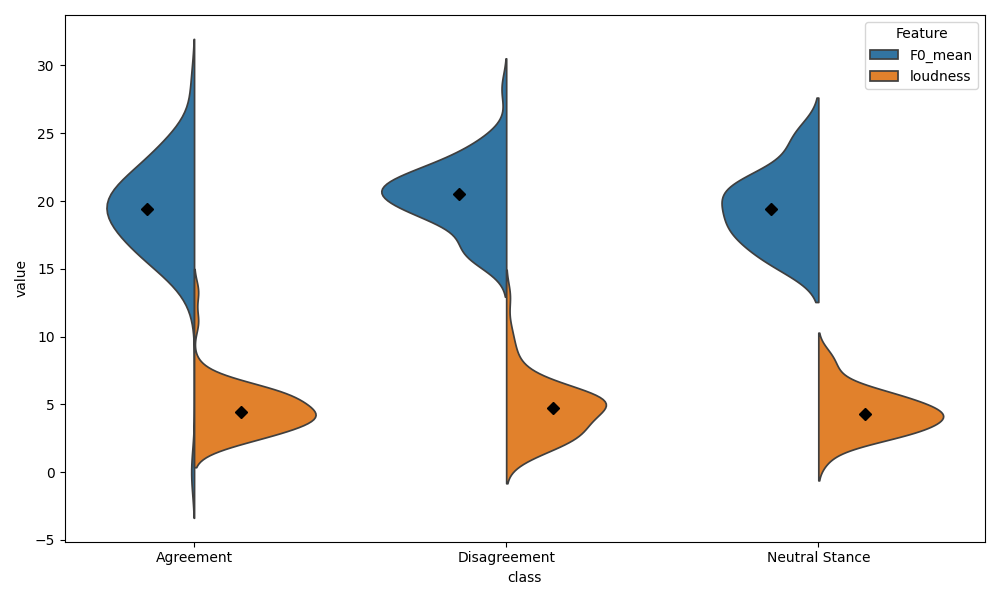}
    \caption{Violin plots of \textit{F0\_mean} and \textit{loudness} across stances. Disagreement shows a higher pitch and louder delivery, Neutral is flatter with lower loudness, while Agreement lies in between.}
    \label{fig:audio_violion}
    \vspace{-3mm}
\end{figure}

To investigate prosodic cues, we extracted acoustic features using openSMILE~\cite{eyben2010opensmile} and visualized the distributions of mean fundamental frequency (F0\_mean) and loudness across stances (Fig.~\ref{fig:audio_violion}). 
The median markers ($\blacklozenge$) highlight subtle but consistent shifts between classes.
The analysis reveals a clear trend in vocal expressiveness: Disagreement is characterized by a higher median pitch and loudness with greater variance; Agreement shows intermediate distributions; and Neutral is the least expressive, with lower loudness and more stable pitch. While significant overlap suggests these features alone are insufficient for robust classification, the distinct prosodic patterns, particularly the heightened vocal intensity in disagreement, confirm their value as supplemental cues within a multimodal framework.

\subsection{Speech Text Analysis}

\begin{figure}[th]
    \centering
    \includegraphics[width=0.99\linewidth]{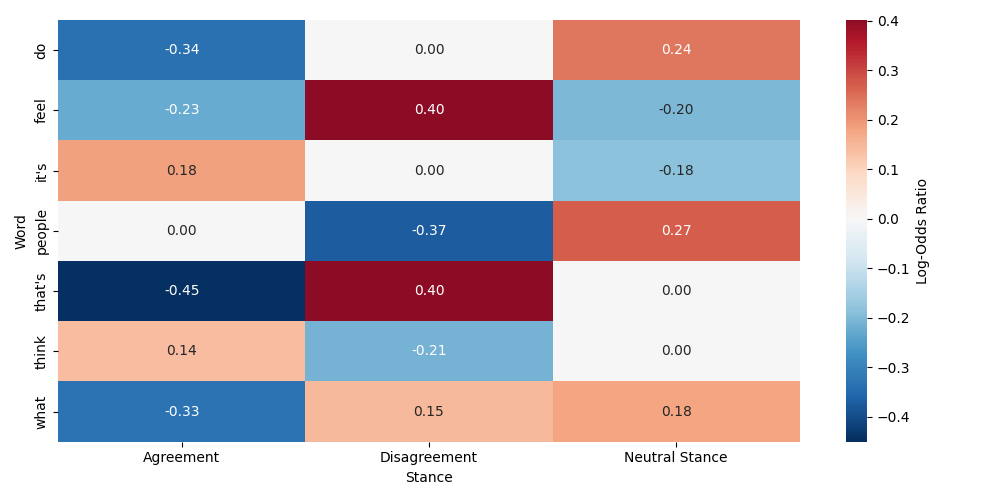}
    \caption{Heatmap of stance–word associations after chi-square filtering}
    \label{fig:text_log-Odds_ratio}
    \vspace{-3mm}
\end{figure}

A fine-grained lexical analysis using chi-square tests and log-odds ratios was conducted to identify stance-sensitive words, with results visualized in Fig.~\ref{fig:text_log-Odds_ratio}. 
The analysis reveals distinct lexical patterns for each stance: disagreement is strongly associated with tokens like ``\textit{that’s}'', ``\textit{people}'', and the subjective marker ``\textit{feel}''; 
agreement is moderately characterized by ``\textit{it’s}'' and ``\textit{think}''; 
and the neutral stance shows a higher usage of interrogative words like ``\textit{what}'' and ``\textit{do}''. 
These findings demonstrate that subtle distributional differences within frequent conversational words can serve as reliable cues for stance detection, supporting the use of stance-specific lexical priors to improve classification models.

To analyze the semantic content of the transcribed speech, we fine-tuned a pre-trained BERT model~\cite{bert} for the three stances classification task (Agree, Disagree, Neutral). 
All sentences are mixed and split into training and test datasets.
Upon evaluation on our hold-out test set, the model achieved an overall accuracy of 63.5\%. 
While this result significantly surpasses the random baseline, a detailed breakdown of performance revealed that the model excelled at identifying Neutral statements but frequently confused instances of Agree and Disagree. This suggests that while the model effectively learned to recognize objective language, the nuanced expressions of agreement and disagreement presented a more substantial challenge.




\subsection{Self-Report Analysis}
\begin{figure}[b]
  \centering
  \begin{subfigure}[t]{0.49\linewidth}
    \includegraphics[width=\textwidth]{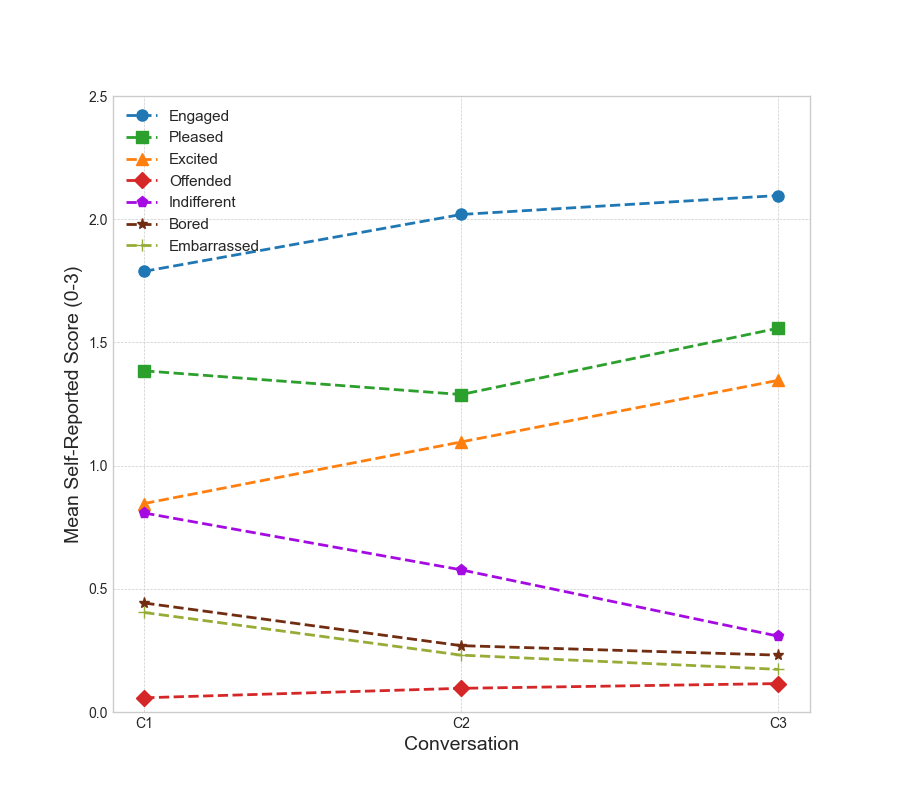}
    \caption{Emotional Trajectory Across Conversations}
  \end{subfigure}
  \hfill
  \begin{subfigure}[t]{0.49\linewidth}
    \includegraphics[width=\textwidth]{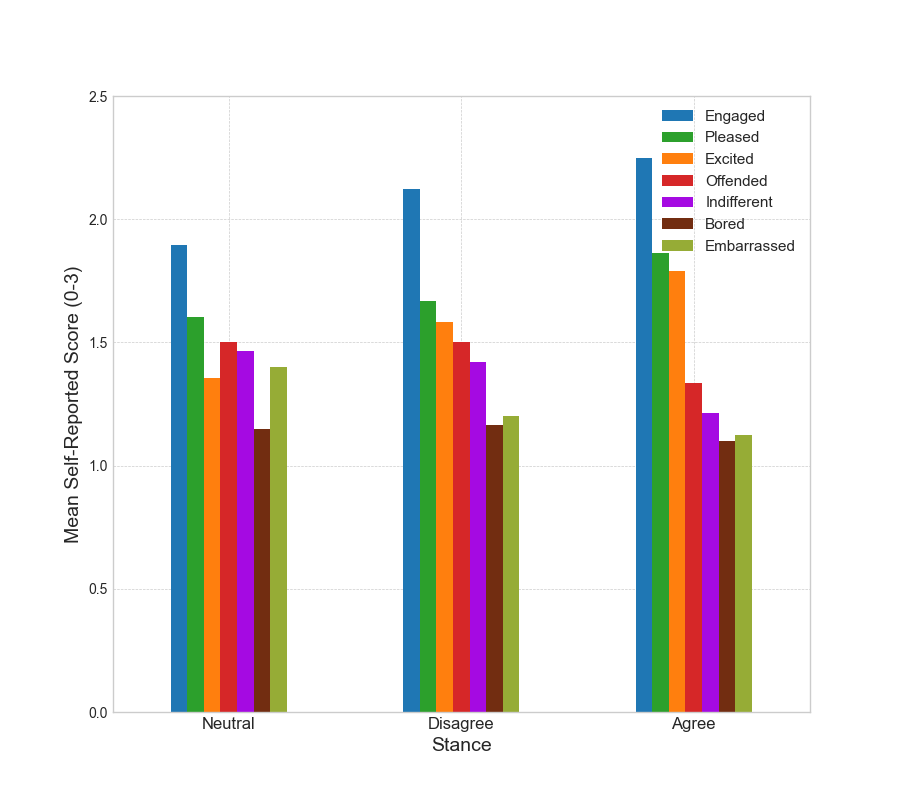}
    \caption{Emotional Profile by Stance}
  \end{subfigure}
  \caption{Self-Reported Emotional Dynamics in Dyadic Conversations}
  \label{fig:self-report}
\end{figure}

Fig.~\ref{fig:self-report} presents the mean self-reported emotional intensity scores, 0 to 3 as None to Intense, revealing a compelling interplay between a temporal "warming up" effect and the intrinsic emotional profile of each stance. 
As conversations progressed from first to last (a), participants reported a significant increase in positive affect (e.g., Engaged, Excited) and a decrease in passive-negative feelings (e.g., Indifferent, Embarrassed), which is consistent with a social familiarization process between strangers.
Complementing this trend, the stance-specific profiles (b) demonstrate that active stances (Agree, Disagree) were markedly more engaging than Neutral ones. 
Critically, the heightened levels of Embarrassed and Indifferent in Neutral conversations can be interpreted as a compound effect, stemming from the emotional passivity of the stance itself, amplified by the initial social friction of being the first interaction in our predominantly fixed-order experimental design.

\begin{figure}[t]
    \centering
    \includegraphics[width=0.95\linewidth]{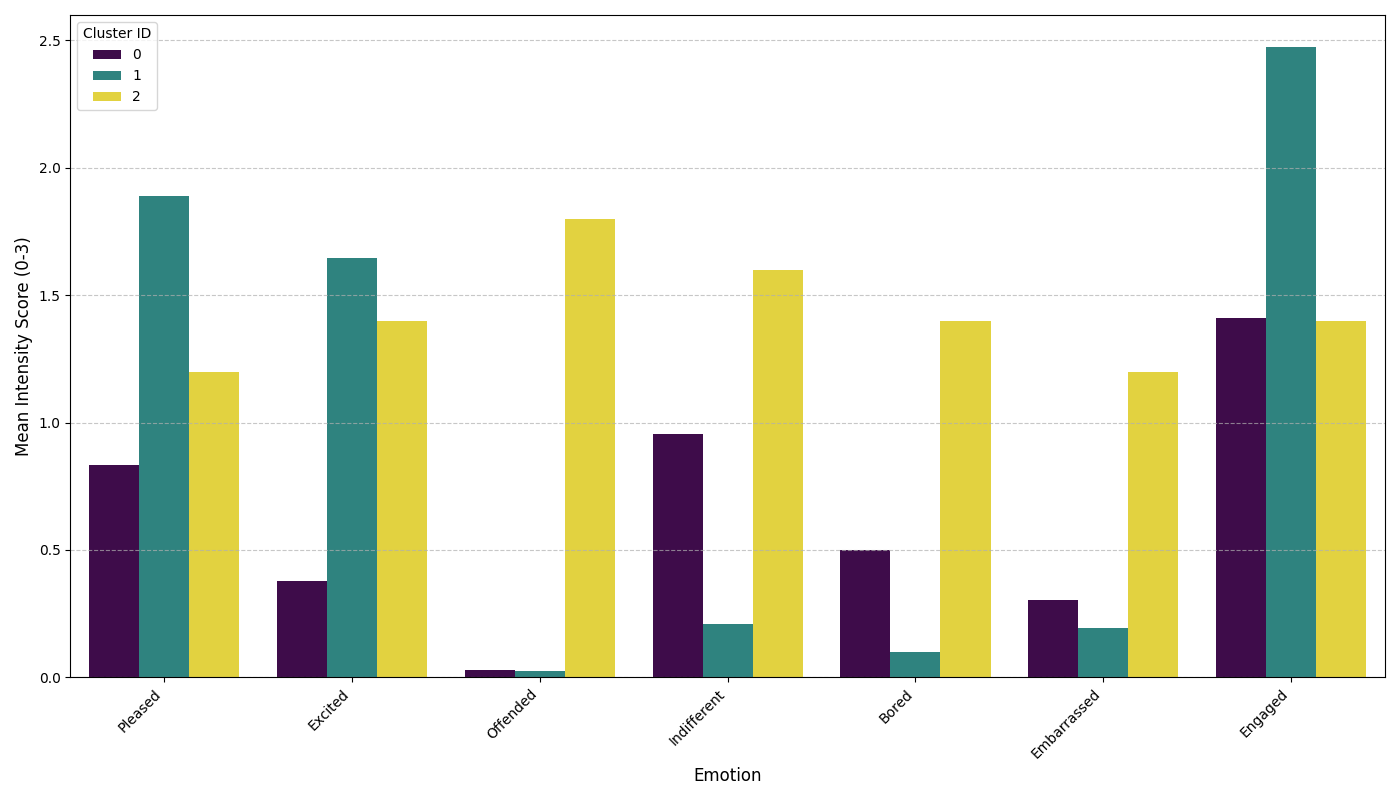}
    \caption{Emotional Profiles of Discovered Self-Report Clusters}
    \label{fig:self_report_kmean}
    \vspace{-3mm}
\end{figure}

Applying K-Means clustering (K=3) to our self-report data partitioned the conversational experiences into three distinct emotional profiles, and the visualization is shown in Fig.~\ref{fig:self_report_kmean}.
The first profile, "Positive-Engaged," is marked by high levels of engagement and positive affect (e.g., Pleased, Excited). 
The second, "Negative-Confrontational," is uniquely characterized by a high intensity of feeling Offended. 
The final "Passive-Detached" profile is defined by a general lack of emotion and low scores across all affective dimensions.
Positive-Engaged, Negative-Confrontational, and Passive-Detached were distributed in 53.6\%, 43.1\%, and 3.3\%, which is extremely unbalanced.
This distribution strongly suggests that the disagreements elicited by our protocol were not emotionally intense; rather, these differences are largely managed within socially acceptable, non-hostile frameworks.

\section{Conclusion and Future Work}

In this work, we introduced Inter-Stance, a large-scale multimodal dataset designed to capture the complex dynamics of conversational stance. 
Our comprehensive analyses serve to validate the dataset's utility by revealing distinct and meaningful patterns across recorded modalities. 
We successfully identified unique affective profiles for Agree, Disagree, and Neutral stances from self-reports; quantified context-dependent facial mimicry; and demonstrated the highly subject-dependent nature of both visual and physiological cues through comparative modeling experiments. 
These findings demonstrate the utility of the Inter-Stance corpus as a resource for studying the nuanced, embodied nature of human social interaction.

Future Work:
First, the analysis of visual cues can be refined by moving from categorical emotion labels to facial action units (AUs).
Second, the inclusion of the audio modality, using low-level acoustic features such as Mel Frequency Cepstral Coefficients (MFCCs), would facilitate the exploration of paralinguistic cues.
Finally, the self-report's emotion, along with distinct emotional profiles discovered through the clustering of self-report data, can be leveraged as ground-truth labels for new supervised learning tasks.


\section{Acknowledgment}
This work is supported by the NSF under grants CNS-1629898, CNS-1629856, CNS-1629716, and the Center of Imaging, Acoustics, and Perception Science (CIAPS) of the Research Foundation of Binghamton University.

{\small
\bibliographystyle{ieee}
\bibliography{egbib}
}

\end{document}